\documentclass{llncs}

\usepackage[utf8]{inputenc}
\usepackage{graphicx}
\usepackage[caption=false]{subfig}

\usepackage{pgfplots}
\usepackage{tikz}
\usetikzlibrary{calc}
\usetikzlibrary{positioning}

\usetikzlibrary{external}
\usepackage{amsmath}
\usepackage{placeins}

\begin{document}
\frontmatter          
\pagestyle{headings}  
%
\mainmatter              
\title{Objectness Scoring and Detection Proposals in Forward-Looking Sonar Images with Convolutional Neural Networks}
\author{Matias Valdenegro-Toro}
\authorrunning{Valdenegro-Toro et al.} 
%
\tocauthor{Matias Valdenegro-Toro}
\institute{Ocean Systems Laboratory\\
School of Engineering \& Physical Sciences\\
Heriot-Watt University, EH14 4AS, Edinburgh, UK\\
Email: m.valdenegro@hw.ac.uk}

\maketitle              

\begin{abstract}
Forward-looking sonar can capture high resolution images of underwater scenes, but their interpretation is complex. Generic object detection in such images has not been solved, specially in cases of small and unknown objects. In comparison, detection proposal algorithms have produced top performing object detectors in real-world color images.
In this work we develop a Convolutional Neural Network that can reliably score objectness of image windows in forward-looking sonar images and by thresholding objectness, we generate detection proposals.
In our dataset of marine garbage objects, we obtain 94 \% recall, generating around 60 proposals per image. The biggest strength of our method is that it can generalize to previously unseen objects. We show this by detecting chain links, walls and a wrench without previous training in such objects.
We strongly believe our method can be used for class-independent object detection, with many real-world applications such as chain following and mine detection.
\end{abstract}

\section{Introduction}

Autonomous Underwater Vehicles (AUVs) are increasingly being used for survey and exploration of underwater environments. For example, the oil and gas industry requires constant monitoring and surveying of seabed equipment, and marine researchers require similar capabilities in order to monitor ocean flora and fauna.

The perception capabilities of AUVs are not comparable to land and air vehicles. Most of the perception tasks, such as object detection and recognition, are done in offline steps instead of online processing inside the vehicle. This limits the applications fields where AUVs are useful, and strongly decreases the level of autonomy that this kind of vehicles can achieve.

Most of these limits on perception capabilities come directly from the underwater environment. Water absorbs and scatters light, which limits the use of optical cameras, specially near coasts and shores due to water turbidity and suspended material. Typical perception sensors for AUV are different kinds of Sonar, which uses acoustic waves to sense and image the environment. Acoustic waves can travel great distances on water with small attenuation, depending on frequency, but interpreting an image produced by a sonar can be challenging.

One type of sonar sensor is Forward-Looking Sonar (FLS), where the sensor's field of view looks forward, similar to an optical camera. Other kinds of sonar sensors have downward looking fields of view in order to survey the seabed. This kind of sensor is appropriate for object detection recognition in AUVs.

Object detection in sonar imagery is as challenging as other kinds of images. Methods from the Computer Vision community have been applied to this kind of images, but these kind of methods only produce class-specific object detectors. Most research has been performed on detecting marine mines \cite{reed2004automated}, but constructing a class-agnostic object detector is more useful and will greatly benefit AUV perception capabilities.

Computer Vision literature contains many generic object detection algorithms, called detection proposals \cite{alexe2012measuring} \cite{hosang2015makes}, but these techniques were developed for color images produced by optical cameras, and color-based techniques fail to generate correct proposals in sonar images.
Convolutional and Deep Neural Networks are the state of the art for many computer vision tasks, such as object recognition \cite{krizhevsky2012imagenet}, and they have also been used to generate detection proposals with great success \cite{girshick2015fast} \cite{ren2015faster} \cite{ren2015object}.

The purpose of our work is to build an algorithm for detection proposal generation in FLS images, but our technique can still be used for other kinds of sonar images. Instead of engineering features that are commonly used for object detection, we propose to use a Convolutional Neural Network (CNN) to learn objectness directly from labeled data. This approach is much simpler and we believe that has better generalization performance than other object detection approaches.

\section{Related Work}

Detection Proposals \cite{endres2010category} are class-agnostic object detectors. The basic idea is to extract all object bounding boxes from an image, and compute an objectness score \cite{alexe2012measuring} that can be used to rank and determine interesting objects, with the purpose of posterior classification.

Many methods to extract detection proposals in color images exist.
Rathu et al. \cite{rahtu2011learning} uses cascade of objectness features to detect category-independent objects. Alexe et al. \cite{alexe2012measuring} use different cues to score objectness, such as saliency, color contrast and edge density. 

Selective search by Uijlings et al. \cite{uijlings2013selective} uses a large number of engineered features and superpixel segmentation to generate proposals in color images, which achieves a 99\% recall on many datasets. Girshick et al. \cite{girshick2014rich} combine Selective Search with a CNN image classifier to detect and recognize objects in a common pipeline.

Zitnick et al. \cite{zitnick2014edge} use edge information to score proposals from a sliding window in a color image. Kang et al. \cite{kang2015data} use a data driver approach where regions are matched over a large annotated dataset and objectness is computed from segment properties.

Kuo et al. \cite{kuo2015deepbox} shows how to learn objectness with a CNN with the purpose of reranking proposals generated by EdgeBoxes \cite{zitnick2014edge}, with improved detection performance. A good extensive evaluation of many proposal algorithms is Hosang et al. \cite{hosang2015makes}.

More recent proposal approaches also use CNNs, such as Fast R-CNN \cite{girshick2015fast} and Faster R-CNN \cite{ren2015faster}. Fast R-CNN uses bounding box regression trained over a convolutional feature map that can be shared and used for both detection and classification, but still using initial Selective Search proposals \cite{uijlings2013selective}, while Faster R-CNN uses region proposal networks to predict proposals and objectness directly from the input image, while sharing layers with a classifier and bounding box regressor in a similar way that of Fast R-CNN.

Object detection in sonar images is mostly done with several kinds of engineered features over sliding windows and a machine learning classifier \cite{reed2004automated} \cite{isaacs2015sonar} \cite{williams2011fast} \cite{fandos2011optimal}, template matching \cite{myers2010template} \cite{hurtos2013automatic} is also very popular, as well as computer vision techniques like boosted cascade of weak classifiers \cite{sawas2010cascade}. In all cases this type of approach only produces class-specific detectors, where generalization outside of the training set is poor.

While proposal methods are been successful on computer vision tasks, color image features are not appropriate for sonar images, due to the different interpretation of the image content. Some methods such as EdgeBoxes \cite{zitnick2014edge} could be applied to sonar images, but it is well known that edges are unreliable in this kind of images due to noise and point of view dependence.

\section{Forward-Looking Sonar Imaging}

A Forward-Looking Sonar is an acoustic sensor that is similar to an optical camera \cite{hurtos2013automatic}, but with two major differences: Sound waves are emitted and the acoustic return is analyzed to discover scene structure, and the output image is similar to a top view of the scene instead of the typical front view of a optical camera. An FLS that uses high-frequency acoustic pulses can capture high resolution images of underwater scenes at distances ranging from 1 to 10 meters at frame rates of up to 15 Hz. This kind of device is used for survey and object identification in underwater environments with AUVs \cite{negahdaripour2005processing}.

An FLS has a fan-shaped field of view, with fixed viewing angles and configurable distances. The output of most sonar sensors is a one channel image that represents the amount of acoustic return from the scene, usually sampled along the sensor's field of view. A typical characteristics of acoustic images are shadow (dark) and highlight (light) regions, produced when objects block and reflect sound waves. The length of shadow regions depends on the height of the object. Ghost reflections are produced by spurious acoustic return from undesired sources, such as walls, the water surface, large material changes between objects and interference inside small objects. The fan-shaped field of view introduces pixel shape distortions, since the size of a pixel in the image now depends on the distance from the origin. Farther distances map to bigger pixels that have increasing uncertainty, while closer distances have smaller uncertainties \cite{negahdaripour2005processing}.

\begin{figure}
    \centering
    \subfloat[FLS Image containing a Drink Carton, Can and Bottles]{
        \includegraphics[width=0.4\textwidth]{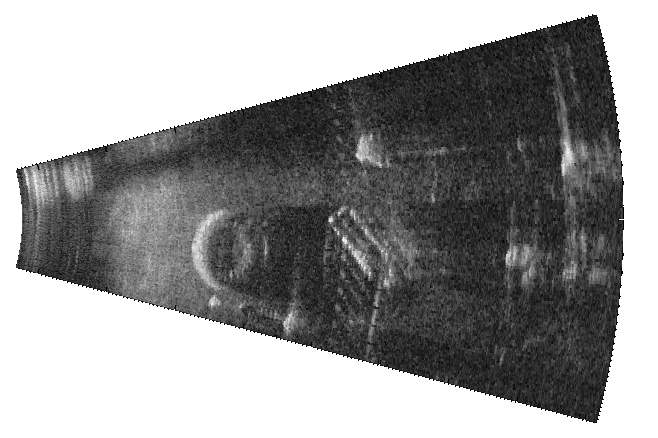}}
    \subfloat[Tin Can and Wall, note the reflections in both objects]{
        \includegraphics[width=0.3\textwidth]{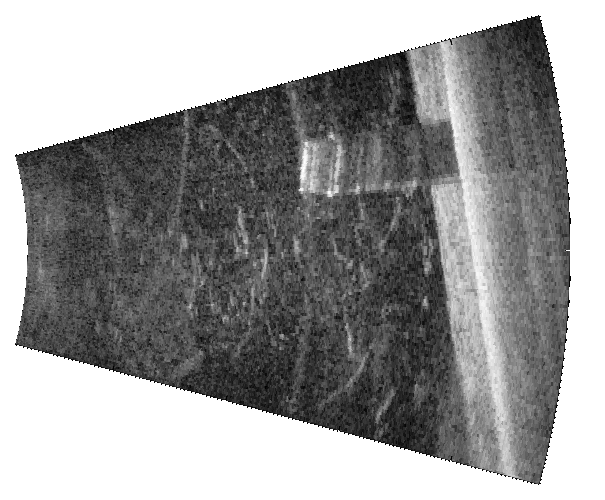}}
    \caption{Sample FLS images from our dataset, captured with a ARIS Explorer 3000 sonar.} 
    \label{sampleFLSImages}
\end{figure}

These features make FLS images hard to interpret. One important feature of sonar sensors is that in most cases distance and bearing to the object can be easily recovered directly from sonar data, but elevation information is lost. Fig \ref{sampleFLSImages} shows two FLS images, where each pixel represents approximately 3 mm, and some objects can be seen clearly (The tire and bottle in Fig \ref{sampleFLSImages}a), even with some fine details like the seams on the tire. Fig \ref{sampleFLSImages}b shows typical sonar reflections.

\section{Detection Proposals on FLS Imagery}

Our proposed technique is similar in spirit to \cite{alexe2012measuring} and \cite{kuo2015deepbox}. We propose to use a CNN to learn objectness scores of windows in a FLS image. We slide a $w \times w$ window over the image with a stride of $s$ pixels, but only consider windows that are inside the FLS's field of view. Our technique only requires a image dataset with labeled rectangles representing objects in the image. The objectness scores are estimated directly from the rectangles.\\

Each window is scored by our CNN and objectness is thresholded to select windows that contain objects. Different objectness threshold values $T_o$ will produce varying numbers of proposals. Control over the number of proposals is a desirable property of an proposal algorithm \cite{hosang2015makes}.
Windows with a low objectness score are discarded. Given a labeled dataset containing objects of interest, we construct a training set by running a sliding window over each image and cropping each window that has an  intersection-over-union (IoU) score above a threshold. Given two rectangles $A$ and $B$, the IoU score is defined as:

\begin{equation}
    \text{IoU}(A, B) = \frac{\text{area}(A \cap B)}{\text{area}(A \cup B)}
\end{equation}

The IoU score is commonly used by the computer vision community to evaluate object detection algorithms \cite{zitnick2014edge} \cite{girshick2015fast}.\\

For each cropped window, we estimate the ground truth objectness of that sub-image as a score based on the maximum IoU with ground truth. Our intuition for such scoring is that non-object windows have a very low IoU with ground truth, while object windows have a high IoU with ground truth, but multiple windows might contain the object. IoU decreases as the window moves farther from the ground truth, and we want the same behavior from our objectness scores. This technique also doubles as a data augmentation step, as many training windows will be generated for a single ground truth object. In practice we obtained up to 35 crops from one object.

The ground truth objectness for a window is computed as:

\begin{equation}
    \text{objectness}(\text{iou}) = 
    \begin{cases} 
        1.0 		& \text{if iou} \geq 0.8 \\
        \text{iou}  & \text{if } 0.2 < \text{iou} < 0.8\\
        0.0      	& \text{if } \text{iou} \leq 0.2
    \end{cases}
    \label{iouObjectness}
\end{equation}

Eq \ref{iouObjectness} represents our desired behavior for the ground truth objectness score. Windows with a small IoU are very unlikely to contain an object, and this is reflected as zero objectness, while windows with a high IoU contain objects and get a objectness equals to one. Windows with IoU values in between are assigned objectness scores equals to the IoU.\\

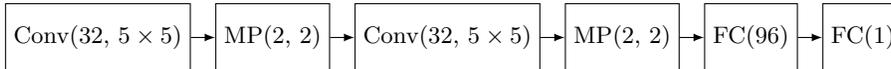
\begin{figure}[!htb]
    \centering
    \begin{tikzpicture}[style={align=center, minimum height=0.9cm}]
    \node[draw](B) {Conv($32$, $5 \times 5$)};
    \node[draw, right=1em of B] (C) {MP(2, 2)};
    \node[draw, right=1em of C] (D) {Conv($32$, $5 \times 5$)};
    \node[draw, right=1em of D] (E) {MP(2, 2)};
    \node[draw, right=1em of E] (F) {FC(96)};
    \node[draw, right=1em of F] (G) {FC(1)};
    \draw[-latex] (B) -- (C);
    \draw[-latex] (C) -- (D);
    \draw[-latex] (D) -- (E);
    \draw[-latex] (E) -- (F);
    \draw[-latex] (F) -- (G);
    \end{tikzpicture}
    \caption{Our Neural Network Architecture}
    \label{arisNetArchitecture}
\end{figure}

Our CNN architecture is shown in Fig \ref{arisNetArchitecture}. It is a 4 layer network that takes $96 \times 96$ pixel images as input, and has $1.4$ million trainable parameters. The first layer convolves the input image with 32 $5 \times 5$ filters, then applies $2 \times 2$ Max-Pooling (MP), then the same process is repeated by another convolution layer with 32 $5 \times 5$ filters and $2 \times 2$ Max-Pooling. The classifier layers are one Fully Connected layer (FC) with 96 neurons, and another FC layer with one output neuron. All layers use the ReLU non-linearity except the last layer, which uses a sigmoid function to output objectness scores in the $[0, 1]$ range.

Our architecture was initially modeled to be similar to LeNet \cite{lecun1998gradient}, by stacking Convolution, ReLU and Max-Pooling blocks for convolutional feature learning, and then produce objectness scores with a fully connected layer. This initial architecture worked best and generalizes very well over unseen data and out of sample images. Removing any layer decreases recall, and stacking more Conv-ReLu-MaxPool or adding fully connected layers leads to overfitting. We believe that this is an appropriate architecture for the problem and for the amount of data that we possess.

Batch Normalization \cite{ioffe2015batch} layers are inserted after each Max-Pooling layer, and after the first FC layer, in order to prevent overfitting and accelerate training. We also tested Dropout \cite{srivastava2014dropout} but Batch Normalization gave better generalization performance. We trained our network with the standard mean square error (MSE) loss function and mini batch gradient descent, with a batch size of 32 images, and used the ADAM optimizer \cite{kingma2014adam} with a initial learning rate $\alpha = 0.1$. We train for 10 epochs, and do early stopping when the validation loss stops improving (normally after 5 epochs). Our network is implemented in Python with Keras and Theano.\\

The dataset used to train this network is generated as follows. A $96 \times 96$ sliding window with stride of $s = 8$ pixels is run over each image in the dataset, and all windows that have IoU $\geq 0.5$ with ground truth are added to the training set, and objectness labels are determined with Eq \ref{iouObjectness}. This generates a variable number of crops for each image, up to 35 window crops. Some objects generated no crops, due to having IoU with ground truth less than 0.5, but were still included in the dataset to keep examples with intermediate objectness values.
Negative samples were generated by selecting 20 random windows from the same sliding window with a maximum IoU $\le 0.1 $ with ground truth. Each negative crop was assigned objectness label of zero.

A common issue is that many intersecting proposals are present in the output, due to the use of a sliding window. These can be reduced by performing non-maxima suppression over intersecting proposals with a minimum IoU threshold.

\section{Experimental Evaluation}

In this section we evaluate our proposed approach. We captured 2500 FLS images with an ARIS Explorer 3000 sonar, containing different objects that are interesting for the underwater domain, such as garbage objects, a model valve, a small tire, a small chain, and a propeller. We performed a 70\%/30\% split of the image set, where the training set contains 1750 images, and the test set 750 images. The training set was 85\%/15\% split into train and validation sets, the latter set for purpose of early stopping. We introduce up-down and left-right flips of each training image for data augmentation.\\


To evaluate at test time, we run a sliding window over the image, with the same parameters used to generate the training set, and threshold the predicted objectness score from our CNN. Any window with an objectness score bigger than $T_o$ is output as a object proposal. $T_o$ is a parameter that can be tuned by the operator to decide the number of proposals to generate. A detection is considered correct if it has IoU $\geq 0.5$ with ground truth.

\begin{figure}[!htb]
    \begin{tikzpicture}
        \begin{axis}[width = 0.48\textwidth, xlabel={Objectness Threshold ($T_o$)}, ylabel={Test Recall}, xmin=0, xmax=1.0, ymin=0, ymax=1.0, ymajorgrids=true, xmajorgrids=true, grid style=dashed, legend pos = south west]        
            \addplot[color=blue, mark=triangle] table[x  = threshold, y  = recall, col sep = space] {data/thresholdVsRecallAtIoU0.5.csv};
            \addlegendentry{Our Method}
            \addplot[color=gray] table[x  = threshold, y  = recall, col sep = space] {data/thresholdVsRecallRandom.csv};
            \addlegendentry{Baseline}
        \end{axis}        
    \end{tikzpicture}
    \begin{tikzpicture}
        \begin{axis}[width = 0.48\textwidth, xlabel={Objectness Threshold ($T_o$)}, ylabel={Number of Proposals}, xmin=0, xmax=1.0, ymin=0, ymajorgrids=true, xmajorgrids=true, grid style=dashed,legend pos=south west]
            \addplot[color=red, mark=x] table[x  = threshold, y  = numberOfProposals, col sep = space] {data/thresholdVsRecallAtIoU0.5.csv}; 
            \addplot[color=gray] table[x  = threshold, y  = numberOfProposals, col sep = space] {data/thresholdVsRecallRandom.csv};    
        \end{axis}
    \end{tikzpicture}
    \caption{Objectness Threshold versus Recall and Number of Proposals over our test set at 0.5 IoU with ground truth. Blue is recall, red is number of generated proposals, and gray is the baseline. Best viewed in color.}
    \label{objectnessVsRecallAndNumberOfProposals}
\end{figure}
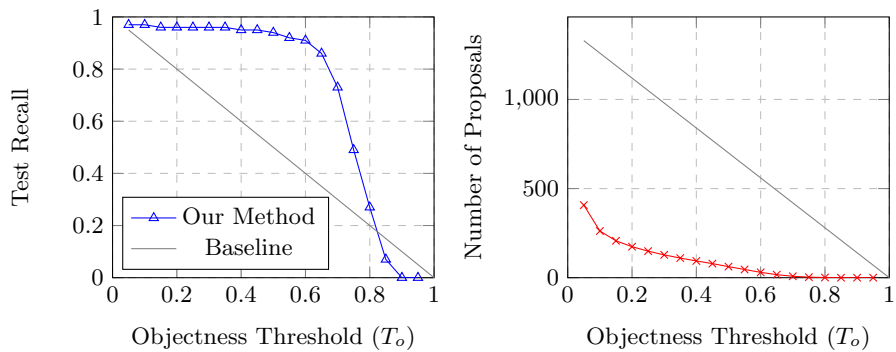

Fig \ref{objectnessVsRecallAndNumberOfProposals} presents our main quantitative results. Since there are no other comparable algorithms that generate proposals in FLS images, we defined our baseline as random scoring of image windows, with a uniform distribution in $[0, 1]$. Our method is clearly superior than the baseline, except when the objectness threshold is high. We believe that our results show that our proposed technique is promising. At $T_o = 0.5$, we obtain $94 \%$ recall, with an average of $62 \pm 33$ generated proposals. The number of proposals is usually a weak indicator of the detector quality, since producing a large number of proposals will score a high recall but will be computationally expensive to evaluate such amount of proposals. A good detection proposal algorithm will generate a small amount of proposals that only cover objects. For comparison, the number of total windows to evaluated in a FLS image is around 1400.\\

Most missed detections correspond to inaccurate localization, due to the sliding window approach, specially for small objects. A large fraction of predicted scores are less than 0.75, which can be seen as the number of proposals drops to zero at $T_o = 0.8$, this show a bias in the learned network. We expected objectness scores to be more evenly distributed in the [0, 1] range.\\

\begin{figure}[!htb]
    \centering
    \subfloat[]{
        \includegraphics[width=0.33\textwidth]{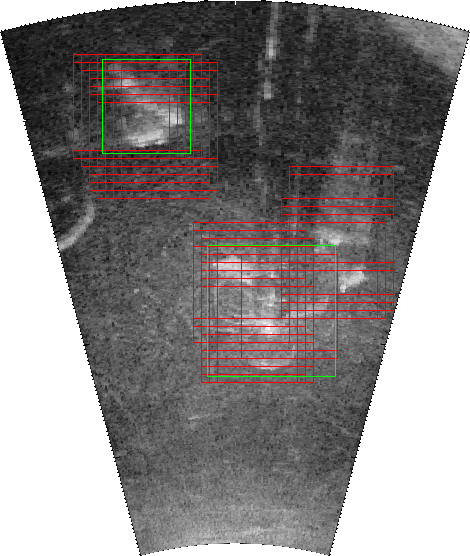}}
    \subfloat[]{
        \includegraphics[width=0.260\textwidth]{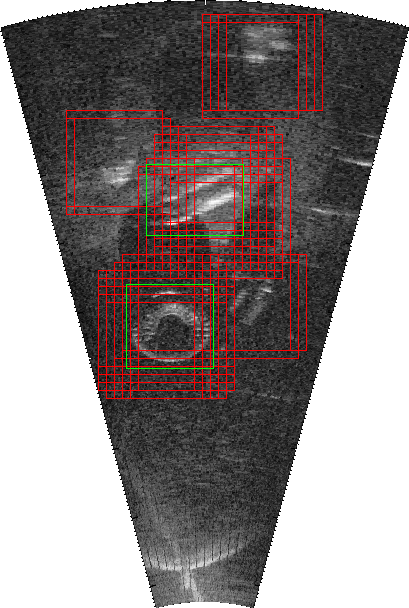}}
    \subfloat[]{
        \includegraphics[width=0.260\textwidth]{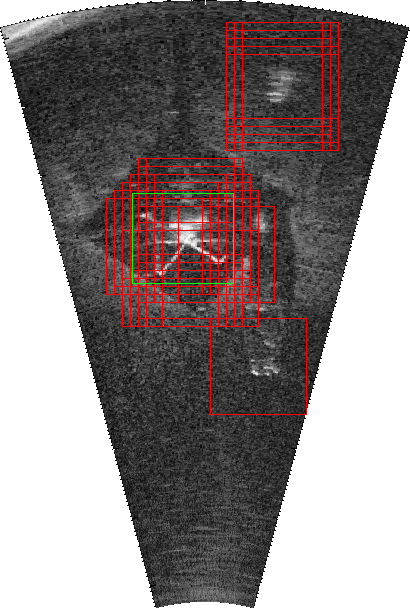}}

    \subfloat[]{
        \includegraphics[width=0.33\textwidth]{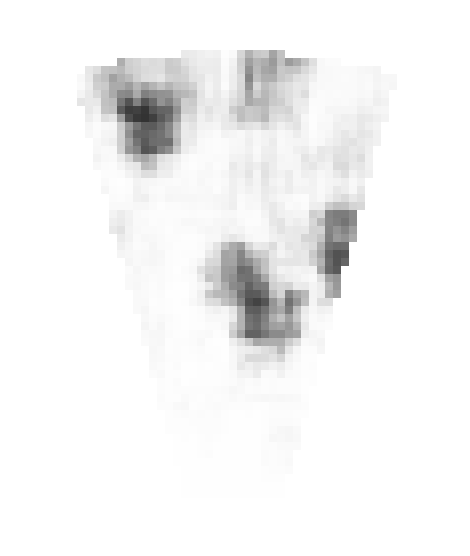}}
    \subfloat[]{
        \includegraphics[width=0.260\textwidth]{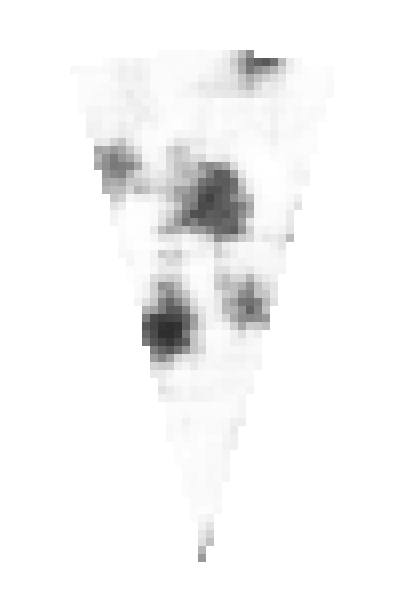}}
    \subfloat[]{
        \includegraphics[width=0.260\textwidth]{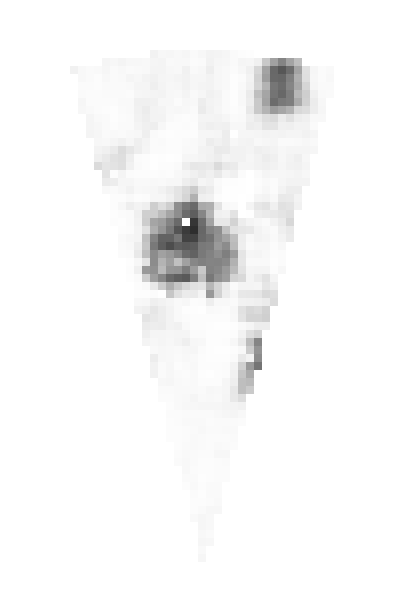}}    
    
    \caption{Results on test images. Top row shows proposals generated with $T_o = 0.5$. Red rectangles are our proposals, while green is ground truth. Note how proposals cover all objects, even unlabeled ones. Bottom row shows heatmaps. White represents low objectness, and black is high objectness scores. Best viewed in color.} 
    \label{testDetections}
\end{figure}

Fig \ref{testDetections}[a, b, c] shows sample detections over our test images. Proposals have a good cover of objects in the image, and surprisingly they also cover objects that are not labeled, which shows that the generalization performance of our approach is good. Multiple proposals cover each object, indicating a poor localization accuracy, but it can be improved by using non-maxima suppression.

Increasing $T_o$ will decrease the number of proposals, producing more accurate detections, but skipping untrained objects. Typically objects that are not seen during training receive lower objectness scores when compared to trained objects.

As qualitative results, we also generated a heatmap of scores produced by the network over the image. This is done by scoring each rectangle in the sliding window, and drawing a $s \times s$ rectangle on the proposal's center. This produces empty zones at edges of the sonar's field of view. Heatmaps are shown in Fig \ref{testDetections}[d, e, f].

The heatmaps show that high objectness scores are only produced over objects and not over background, which validates that our technique works appropriately.\\

\begin{figure}[htb]
    \centering
    \subfloat[Chain]{
        \includegraphics[width=0.30\textwidth]{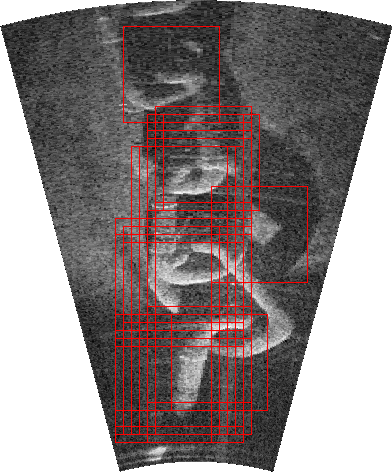}}
    \subfloat[Wall Reflections]{
        \includegraphics[width=0.281\textwidth]{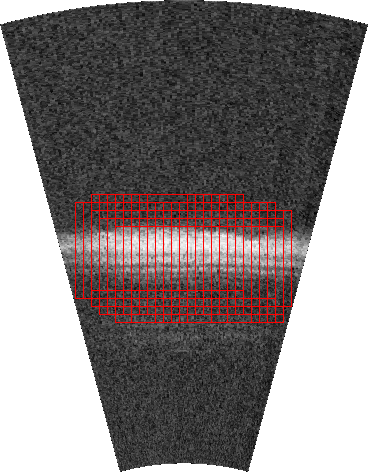}}
    \subfloat[Wrench]{
        \includegraphics[width=0.242\textwidth]{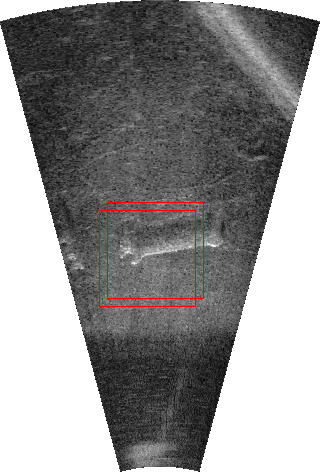}}
    
    \subfloat[Chain]{
        \includegraphics[width=0.30\textwidth]{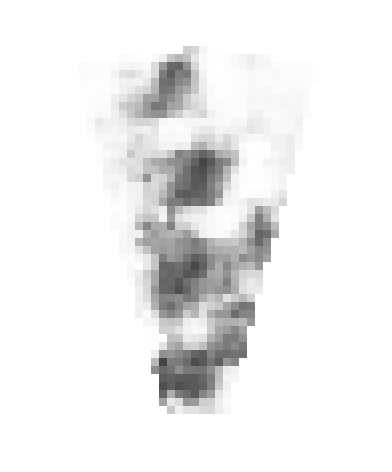}}
    \subfloat[Wall Reflections]{
        \includegraphics[width=0.281\textwidth]{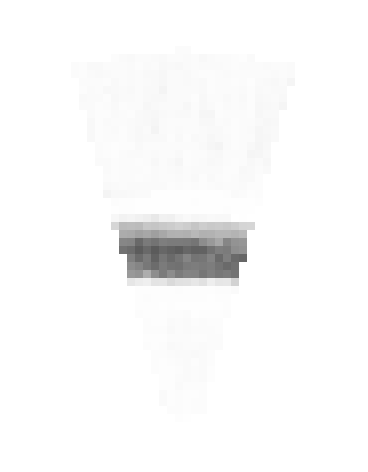}}
    \subfloat[Wrench]{
        \includegraphics[width=0.242\textwidth]{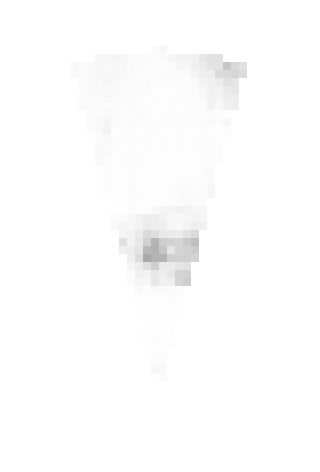}}
            
    \caption{Results on out-of-sample images. Top row shows proposals generated at $T_o = 0.5$. Our training set does not contain any object present in these images, but yet our network can generate very good proposals on them. Bottom row shows heatmaps. White represents low objectness, and black is high objectness scores. Best viewed in color.} 
    \label{outOfSampleDetections}
\end{figure}

We also evaluated our network's generalization abilities outside of the training set. We obtained images of objects that are not present in the training set, like a Chain (provided by the University of Girona), a Wall, and a Wrench. The proposals generated on such images are shown in Fig \ref{outOfSampleDetections}[a, b, c]. We expected that our technique would generate a low amount of proposals with very low position accuracy, but results show that the network can successfully generate proposals over object that it has not seen during training. We also provide heatmaps over these objects in Fig \ref{outOfSampleDetections}[d, e, f].

\FloatBarrier
\section{Conclusions and Future Work}

In this work we have presented a CNN approach at objectness estimation and detection proposal generation for FLS images. Our approach is simple, only requiring images with labeled objects, and works surprisingly well, specially in objects that are not present in the training set, which is a very desirable property of any object detector.

Over our dataset we obtain a $94 \%$ recall at objectness threshold of 0.5, and by lowering the threshold to 0.1 we obtain $97 \%$ recall.\\

Neural network-based methods are responsible for large breakthroughs in computer vision for color images, and we believe that similar improvements are possible in other domains, such as FLS images that we used in this work.
Detection proposal approaches imply that a generic object detector can be shared by many different perception tasks, while only a recognition stage has to be developed for specific kinds of objects. We believe that our approach will be useful to improve AUV perception capabilities, as well as increasing their autonomy.\\

Still there is much work to be done in this field. Our method is slow, taking 12 seconds per frame. We believe that computation time can be improved by converting our network into a Fully Convolutional Neural Network \cite{long2015fully}.

Our method only uses a single scale, and introducing multiple scales could improve detection results, specially for smaller objects. We also would like to combine our approach with a object recognition system for integrated detection and classification in FLS images.

\section*{Acknowledgements}

This work has been partially supported by the FP7-PEOPLE-2013-ITN project ROBOCADEMY (Ref 608096) funded by the European Commission. We thank CIRS at the University of Girona for providing images from their chain dataset, Leonard McLean for his help in collecting the data used in this paper, and Polett Escanilla for her assistance in preparing the figures in this paper.

%
%

\bibliographystyle{splncs03}
\bibliography{proposal-fls-bibliography}

\end{document}